\documentclass[runningheads]{llncs}

\usepackage{subfiles} 
\usepackage{graphicx}
\usepackage{verbatim} 
\usepackage{circuitikz} 
\usepackage{graphicx} 
\usepackage{algorithm} 
\usepackage{algorithmic} 
\usepackage{amsmath} 

\begin{document}

\title{\textbf{Integrate-and-Fire Neurons for Low-Powered Pattern Recognition}}
\titlerunning{IF Neuron for Low-Powered Pattern Recognition}

\author{Florian Bacho \and Dominique Chu}

\institute{CEMS, School of Computing, University of Kent, Canterbury CT2 7NF, UK
\email{fb320@kent.ac.uk}}
	
\maketitle

\begin{abstract}
Embedded systems acquire information about the real world from sensors and process it to make decisions and/or for transmission. In some situations, the relationship between the data and the decision is complex and/or the amount of data to transmit is large (e.g. in biologgers). Artificial Neural Networks (ANNs) can efficiently detect patterns in the input data which makes them suitable for decision making or compression of information for data transmission. However, ANNs require a substantial amount of energy which reduces the lifetime of battery-powered devices. Therefore, the use of Spiking Neural Networks can improve such systems by providing a way to efficiently process sensory data without being too energy-consuming. In this work, we introduce a low-powered neuron model called Integrate-and-Fire which exploits the charge and discharge properties of the capacitor. Using parallel and series RC circuits, we developed a trainable neuron model that can be expressed in a recurrent form. Finally, we trained its simulation with an artificially generated dataset of dog postures and implemented it as hardware that showed promising energetic properties.

\end{abstract}

\section{Introduction}

Embedded systems acquire physical measurements of the real world from sensors before performing simple computations \cite{pattern_recognition_embedded_systems}. From signal acquisition, these systems often require a transformation of the data to make decisions or compress the information for transmission. Pattern recognition is an important area in the emergence of intelligent systems the classification of patterns from sensory information into categories is necessary to achieve a goal \cite{pattern_recognition_embedded_systems}. For example, recent years have seen the development of new animal-attached devices called \textit{Biologgers} which are used to monitor the environment, track locations and quantify the behaviour of certain species \cite{biomarkers_for_animal_health}. These devices sometimes use transmission technologies such as Very High Frequency (VHF), acoustic telemetry or, more recently, orbiting satellites to monitor certain species over a long period. However, data transmission has a high cost not only financially, but also in terms of energy. This can be problematic on battery-powered devices. Thus, to optimise the lifetime of remote devices, the number of transmissions must be minimized. As some sensors often run at a high sampling frequency – typically between 10Hz and 1000hz for inertial sensors – the amount of collected data becomes so large that transmission becomes difficult without any compression or processing. To reduce this amount of data, embedded classifiers can directly process the sensor values, which significantly reduces the information to transmit. For example, some methods have been used on biologgers to classify animal activities from inertial data using machine learning approaches, especially Artificial Neural Networks \cite{identification_of_behavior_freely_moving_dogs,using_tri_axial_acceleration_to_identify_behavior,identification_animal_movement_patterns_tri_axial_magnetometry}
\par
Artificial Neural Networks (ANNs) are one of the most powerful methods to solve classification problems. ANNs try to mimic the behaviour of biological neurons to find complex relationships between input signals and desired outputs. However, the computation of these artificial neurons is computationally expensive due to complex operations that require substantial amounts of energy or sometimes the use of Graphics Processing Units (GPUs) which makes them unsuitable for battery-powered devices \cite{optimizing_energy_consumption_snn}. Contrary to the abstracted models used in Deep Learning, Spiking Neural Networks (SNN) are biologically plausible artificial neuron models \cite{SNN_single_neurons_populations_plasticity} that transmit information through discrete electrical signals called spike trains \cite{SNN_single_neurons_populations_plasticity,optimizing_energy_consumption_snn}. Spiking neurons integrate synaptic events only when they occur and fire action potentials when the membrane potential reaches a defined threshold \cite{SNN_single_neurons_populations_plasticity}. This event integration property makes them relatively easy to simulate and can also be implemented as energy-efficient dedicated hardware (called neuromorphic chips) \cite{memristor_emulator_with_stdp,loihi_neuromorphic_processor,digital_spiking_neurmorphic_processor,spinnaker_project}. To the best of our knowledge, there is no hardware implementation of SNNs embedded in small remote devices such as biologgers – mainly because of the size of the current neuromorphic hardware. Therefore, it is necessary to bring new simple and non-energy-consuming solutions for embedded pattern recognition in remote systems.
\par
In this paper, we present a simple neuron circuit that can be used for basic pattern recognition in remote systems. This model developed is the \textit{Integrate and Fire} (IF) which is easily implementable as energy-efficient hardware with low-cost components. It integrates successive currents during different amounts of time – according to the inputs – and exploits the charge and discharge capabilities of capacitors to create a trainable and electronically implementable neuron. The model has both excitatory and inhibitory synapses and we introduce it as a recurrent form which makes it suitable for gradient descent optimisations. This model has been chosen for the simplicity of its hardware implementation and its simulation. To validate it, we trained three neurons to classify dog postures using inclination vectors (calculated from inertial data) and implemented them with electronic components to compare the hardware and its simulation.

\section{Results} \label{section_results}

The capacitor is an electronic passive component that creates a potential difference between two conductive plates, analogous to the difference of electric potential of the biological neuron membrane created by ions that flow in and out of the cell. Therefore, the capacitor is often used in computational models of spiking neural networks to reproduce membrane potentials of the biological neurons. Connected in series or parallel with a resistance, the capacitor forms two circuits with distinct charge and discharge properties respectively called series and parallel Resistor-Capacitor circuits (RC). Thus, the IF neuron is mainly composed of passive components: a capacitor that reproduces the membrane potential and resistors that charge (excite) or discharge (inhibit) the neuron.

\subsection{Series RC circuit for excitatory stimulations} \label{series_rc}
\begin{figure}
\centering
    \begin{circuitikz}[scale=0.7,transform shape] \draw
    (-3,1) node[left]{$V_{\mathrm{in}}$}
    to[R, o-, l=$R_e$] (0,1)
    to[C, l=C] (2,1)
    node[ground] {}
    ;
    \end{circuitikz}
    \caption{Electric diagram of the series Resistor-Capacitor (RC) circuit. The circuit is composed of a voltage supplier $V_{in}$, a resistor $R_e$ that is analog to the excitatory synapses of the neuron and a capacitor $C$ that reproduces the membrane potential.}
    \label{fig:series_rc}
\end{figure}
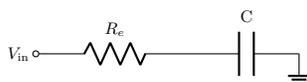
The series RC circuit is defined by a successive resistor $R_e$ which represents the excitatory synapses of the biological neuron and a capacitor $C$ which reproduces the membrane potential – see Figure \ref{fig:series_rc}. Taking into consideration Ohm's law ($I = \frac{V}{R}$), the fact that the current $I_{R_e}$ flowing through the resistor $R_e$ is equal to the current $I_C$ flowing through the capacitor $C$ ($I_{R_e} = I_C$) and that the capacitor component theoretically does not produce any resistance, the current flowing through the circuit depends only on the input voltage $V_{in}$ and the excitatory resistor $R_e$. Thus, the resistor can be seen as a weight defined as $w = \frac{1}{R_e}$ which scales the input value $V_{in}$ such as $I_{R_e} = wV_{in}$. Consequently, the higher the value of the resistor, the lower the current will flow through the capacitor and vice versa. Knowing that the total voltage $V_{in}$ of the circuit is defined as the sum of the voltages $V_R$ and $V_C$ respectively across the resistor and the capacitor ($V_{in} = V_R + V_C$), and the Ohm's law, we can define the following equation:
\begin{equation} \label{series_rc_eq_1}
    \begin{aligned}
    V_{in} = R_eI_{R_e} + V_{C}\\
    \Leftrightarrow I_{R_e} = \frac{V_{in} - V_C}{R_e}
    \end{aligned}
\end{equation}
Equation \ref{series_rc_eq_1} shows that the current flows through the excitatory resistor does not only depend on the input voltage and the resistance but also depends on the voltage across the capacitor. Therefore, the higher the voltage across the capacitor, the lower the current flowing in the circuit will be. To describe the dynamic of the capacitor, the instantaneous rate of voltage change $\frac{dV}{dt}$ of the capacitor is introduced as the current $I$ flowing through the capacitor divided by the capacitance $C$ ($\frac{dV}{dt} = \frac{I}{C}$). Equation \ref{series_rc_eq_1} can be reformulated as:
\begin{equation}\label{series_rc_eq_2}
    \begin{aligned}
    \tau_e\frac{dV_C}{dt} = V_{in} - V_C\\
    \end{aligned}
\end{equation}
where $\tau_e = R_e C$ is the time constant of the series RC circuit which represents the number of seconds needed to reach approximately 63.2\% of the input voltage $V_{in}$ – this value is explained below. For a constant input voltage and a given initial voltage $V_C(t)$ at time $t$, the capacitor voltage $V_C(t + \Delta t)$ after an amount of time $\Delta t$ can be found by integrating equation \ref{series_rc_eq_2}:
\begin{equation} \label{series_rc_eq_3}
    \begin{aligned}
    V_C(t + \Delta t) = V_{in} - (V_{in} - V_C(t)) e^{-\frac{\Delta t}{\tau_e}}
    \end{aligned}
\end{equation}
The fact that the time constant $\tau_e$ represents the amount of time to reach a voltage of approximately 63.2\% of the input voltage is due to of the exponential property of equation \ref{series_rc_eq_3}. Indeed, with an initial voltage of 0, the voltage $V_C(\tau_e)$ reached by the capacitor after a stimulation of $\tau_e$ seconds with an input voltage $V_{in}$ is $V_{in}(1 - e^{-1})$ where $1 - e^{-1} \approx 0.632$.

\subsection{Parallel RC circuit for inhibition}
\begin{figure}
\centering
    \begin{circuitikz}[scale=0.7,transform shape] \draw
    (-2,1) node {}
    to[short](0,1) {}
    to[C, l=C] (0,-3) {}
    to[short](-2,-3) {}
    (-2,-2) node[nigfete] (fet) {}
    (-4,-2.25) node[left]{$V_{in}$}
    to[short, o-] (fet.G) {}
    (-2, 1) to[R, l=$R_i$] (fet.D) {}
    (fet.S) node {}
    to[short](-2,-3) {};
    \end{circuitikz}
    \caption{Electric diagram of the parallel Resistor-Capacitor (RC) circuit controlled by a N-Channel MOSFET transistor.}
    \label{fig:parallel_rc}
\end{figure}
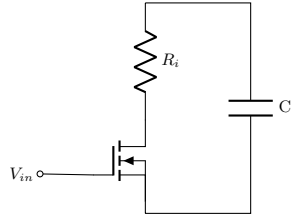

Inhibitory neurons represent 10\% to 20\% of brain population and their activity plays a major role in cognition \cite{from_hiring_to_firing_activation_inhibitory_neurons}. By producing stop signals of excitation and therefore decreasing the membrane potentials of neurons receiving inhibitory stimulus, inhibitory neurons can be seen as regulators of firing rates by maintaining neurons to sub-threshold regimes.
In the IF neuron, an inhibitory connection is implementable with a controlled leakage – similar to the leak of the leaky-integrate-and-fire neuron (LIF). As Figure \ref{fig:parallel_rc} shows, the parallel Resistor-Capacitor (RC) of the LIF neuron circuit can be improved with an N-Channel MOSFET transistor to control the current flowing out of the capacitor. In the parallel RC circuit, the current $I_C$ flowing through the capacitor is equal to the current $I_{R_i}$:
\begin{equation} \label{para_rc_eq_1}
    \begin{aligned}
    I_C=I_{R_i}=\frac{V_C}{R_i}
    \end{aligned}
\end{equation}
Kirchhoff’s voltage law states that the voltage of the capacitor is equal to the voltage drop across the resistor $R_i$ – and the transistor – is equivalent to the voltage $V_C$ of the capacitor:
\begin{equation} \label{para_rc_eq_2}
    \begin{aligned}
    V_{R_i} + V_C = 0\\
    \Leftrightarrow I_{R_i}R_i = -V_C\\
    \Leftrightarrow I_{C}R_i = -V_C
    \end{aligned}
\end{equation}
Finally, as mentioned in section \ref{series_rc}, the instantaneous rate of voltage change $\frac{dV_C}{dt}$ of the capacitor can replace the capacitor's current term in the previous equation:
\begin{equation} \label{para_rc_eq_3}
    \begin{aligned}\tau_{i}\frac{dV_C}{dt} = -V_C
    \end{aligned}
\end{equation}
As for the series RC circuit, the time constant $\tau_i = R_i C$ is introduced which also represents the time required by the discharged capacitor to lose approximately 63.2\% of its voltage. Thus, the previous equation can be integrated to obtain the capacitor's voltage $V_C(t+\Delta t)$ after a stimulation time $\Delta t$:
\begin{equation} \label{para_rc_eq_4}
    \begin{aligned}
    V_C(t + \Delta t) = V_C(t) e^{-\frac{\Delta t}{\tau_i}}
    \end{aligned}
\end{equation}

\subsection{Integrate-and-Fire neuron}
Biological neurons receive several stimuli (excitatory and inhibitory) at their dendrites and having multiple inputs is a necessary condition to allow the IF model to compute separations of multi-dimensional spaces.
Both excitatory and inhibitory can be combined to obtain several inputs. In some specific situations, no excitation is provided by inputs and, for this reason, a bias connection – i.e. a connection always set to 1, as in rate-based models – is introduced to provide a constant stimulation. This allows a permanent charge of the capacitor and the neuron can become excited even if no pattern is provided. In such configuration, the total resistance of parallel resistors is not a simple sum of all the resistances but the inverse of the total resistance is the sum of all inverted resistances ($\frac{1}{R_{total}} = \sum_{i}^{n} \frac{1}{R_i}$). For this reason, computation of the IF model can become complex due to the differences of input stimulation times. For a lack of simplicity, inputs are stimulated one by one and as the capacitor must be charged to allow inhibition of the membrane potential inhibitory stimulations must follow excitatory ones.
Therefore, the inference of the IF model becomes sequential and can be represented under a recurrent form where synapses are stimulated independently.

\subsection{Integrate-and-Fire neuron as a recurrent model} \label{IF_as_recurrent}
\begin{figure}
\centering
\includegraphics[width=0.7\textwidth]{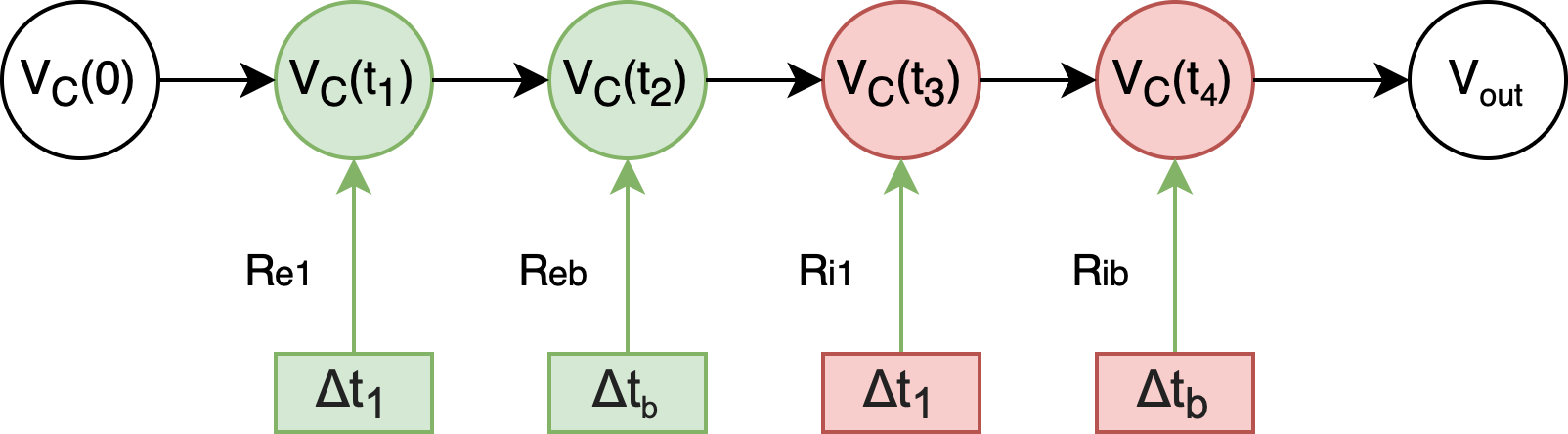}
\caption{Recurrent representation of the inference of the integrate and fire model. Green states represent excitatory stimulations and red states represent inhibition.}
\label{fig:recurrent_model}
\end{figure}
Sequential data are sequences with chronological order. In the deep learning field, this type of data is processed using recurrent units which are feedforward neural networks augmented with the inclusion of internal states of units, introducing a time dimension to the model \cite{critical_review_rnn_sequence_learning}. At each step $t$ of the inference of a recurrent neural network, the states at $t-1$ of the neurons are integrated into the computation. Intrinsically, the integration of stimulus in the IF model depends on the capacitor voltage – see equations  \ref{series_rc_eq_3} and \ref{para_rc_eq_4} – and can be expressed as a recurrent form where each step is a precise synapse stimulation. As presented in Figure \ref{fig:recurrent_model}, each step represents the stimulation of a synapse (excitatory ones first) and the hidden state is the potential of the neuron at time $t-1$ with an initial voltage of 0 – i.e. fully discharged capacitor. Thus, the final hidden state represents the membrane potential of the inferred neuron that can be compared with the voltage threshold to determine if the unit must release a spike or not – this step is achieved by the micro-controller controlling the circuit. The IF model defined as a recurrent form is a continuous and differentiable function which makes it suitable for the gradient descent algorithm. Therefore, some particular set of resistance values makes the IF neuron reach sub-threshold or super-threshold regimes for specific input and this behavior is exploited to achieve classification of patterns. To find the right combinations of resistances, optimisation algorithms can be used such as the well known gradient descent \cite{gradient_descent_for_deep_learning}. The loss of the model can be defined as the Mean Squared Error (MSE) between output membrane potentials and target potentials and the gradient used in the algorithm is computed with respect to resistance values.

\subsection{Dataset, network architecture and training}

To demonstrate our model, we generated an artificial dataset of dog postures and trained a network of three IF neuron on it. It has been generated by using the average inclination vector for each class – i.e. we determined the average tilt of the device for each class – and created many samples by augmenting these vectors with random noise. The tilt of the device can be computed using both accelerometer and gyroscope data from inertial sensors \cite{precise_tilt_angle_detection} which gives a three-dimension vector (pitch, roll and yaw). In this work, three distinct classes of dog postures have been used: stand, sit and lay on the side. The average tilt vectors of each class can be determined by only the pitch and roll axis as following: $\begin{pmatrix}0 & 0\end{pmatrix}$ for stand, $\begin{pmatrix}0 & 0.25\end{pmatrix}$ for sit and $\begin{pmatrix}0.5 & 0\end{pmatrix}$ for lying – the maximum value for each axis is 1. The yaw axis is ignored because it corresponds to the horizontal angle of the device and is irrelevant in this case. From these average tilt vectors, we can generate new input samples with a normally distributed random noise $\epsilon \sim \mathcal{N}(0, 0.04^2)$. 
\par
The model has been implemented as a 3 neuron network – one per class – with both excitatory and inhibitory connections for every input to allow the model to have both types of connection and be flexible enough to achieve correct separations of the input space. The chosen capacitance value for the capacitors is $1e{-6}$ which is small enough to have a low charging time, but high enough to have fine control of the charge, to limit noise and voltage dissipation when implemented as hardware. The maximum stimulation time per input is defined as 50 milliseconds – e.g. an input of value 1.0 will stimulate the corresponding synapse during 50 milliseconds and an input of 0.5 during 25 milliseconds. Finally, the output class is determined by the unit with the highest membrane potential using an argmax operator and the model has been trained using the gradient descent algorithm with a learning rate $\alpha=5e^{-4}$.
\par
During the training, the algorithm did not converge properly due to the scale of resistance values (between $10^3$ and $10^6$) which produces exceedingly large gradients. As the resistances are large and computed into gradients, the scale of gradients also becomes large. This very well known problem is known as \textit{exploding gradient} in machine learning \cite{difficulty_training_recurrent_neural_network}. Many solutions exist to solve the exploding gradient problem such as gradient clipping \cite{difficulty_training_recurrent_neural_network}. However, the gradient clipping method makes gradients too small to converge in an acceptable amount of time – again due to the large scale of resistance values in the model. Another solution has been found to solve the issue: reduce the scale of resistances (between $1^{-3}$ and $1$) and compensate with the capacitance value $C$ of the unit. As the charge and discharge are driven by RC time constants $\tau=RC$, decreasing the resistance $R$ can be balanced by increasing the capacitance $C$. Thus, by scaling down the resistance value, calculated gradients become small enough to obtain stable learning.

\subsection{Weights selection and hardware validation}
\begin{table}
\caption{Resistance values (weights) of stand, lie and sit units. \textit{Excit.} is for \textit{Excitatory} and \textit{Inhib.} is for \textit{Inhibitory}. All values are given in kilohms ($k\Omega$).} \label{model_weights_table}
\begin{tabular}{| c | c c c | c c c |} 
\hline
Output neuron & Excit. x & Excit. y & Excit. bias & Inhib. x & Inhib. y & Inhib. bias \\
\hline\hline
Stand & 20.33 & 101.47 & 1.53 & 9.77 & 6.65 & 1000.00 \\
Lie & 7.61 & 1000.00 & 1000.00 & 1000.00 & 22.44 & 1000.00 \\
Sit & 1000.00 & 5.42 & 1000.00 & 19.57 & 1000.00 & 1000.00 \\
\hline
\end{tabular}
\end{table}
Table \ref{model_weights_table} presents the weights of the model after training. In the IF neuron, a low resistance gives high weight to the input because it lets more current flow in or out of the capacitor and thus has a high contribution in its charge or discharge. Therefore, the contribution of very high resistances is insignificant and can be ignored. For this reason, all resistance values that converged to the maximum resistance ($1000 k\Omega)$ can be ignored in the trained model presented in Table \ref{model_weights_table} and consequently only 9 synapses remain out of the 18.
\par
\begin{figure}
    \centering
     \includegraphics[width=0.7\textwidth]{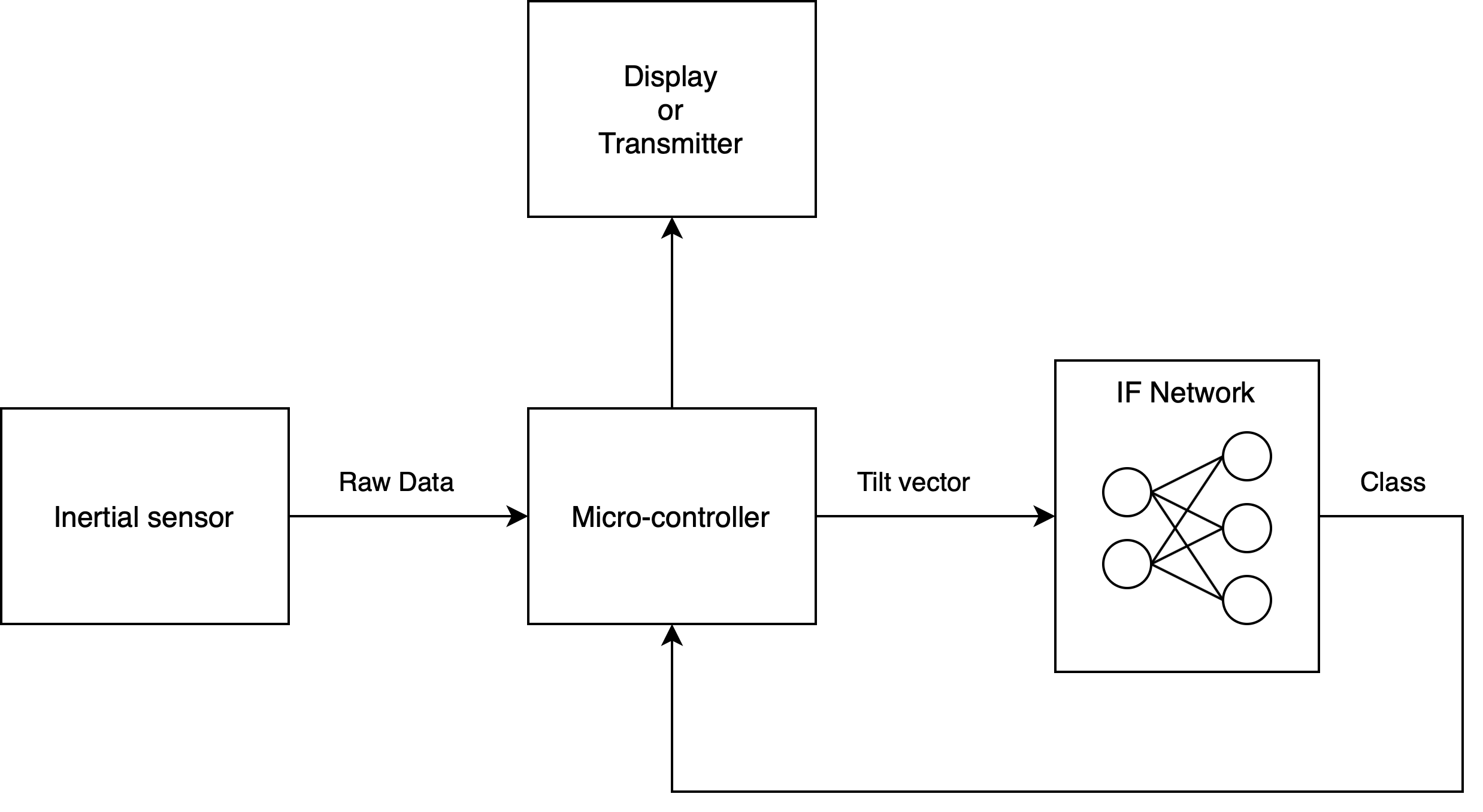}
    \caption{Diagram of the experimental setup. The micro-controller collects the raw accelerometry and gyroscopic data from the inertial sensor, pre-process it to obtain tilt vectors that are sent to a network of IF neuron implemented as hardware. The class inferred by the network can be read by the micro-controller before being sent to the serial display (or a transmitter in real situations of remote systems).}
    \label{fig:setup_diagram}
\end{figure}
After a weight selection (i.e. removing weights that converged to the maximum value), the three IF units for dog posture classification have been implemented as hardware to validate the training. The microcontroller used in this work is an ATmega328P on an Arduino Uno to ease its programming. An inertial unit (MPU-6050) is used to obtain accelerometry and gyroscopic data. Therefore the accelerometry data is used by the microcontroller to determine the gravity vector and the gyroscope data is integrated and combined with the previously computed vector to obtain the precise orientation of the device. Then, the microcontroller stimulates the synapses one by one during variable times depending on the pitch and roll of the device. The synapses charge (excitatory) then discharge (inhibitory) the capacitors using the digital pins. Finally, the microcontroller can read the membrane potential of each neuron by reading the voltage of the capacitors. See Figure \ref{fig:setup_diagram} for a diagram of the setup.
\par
\begin{figure}
    \centering
    \includegraphics[width=0.7\linewidth]{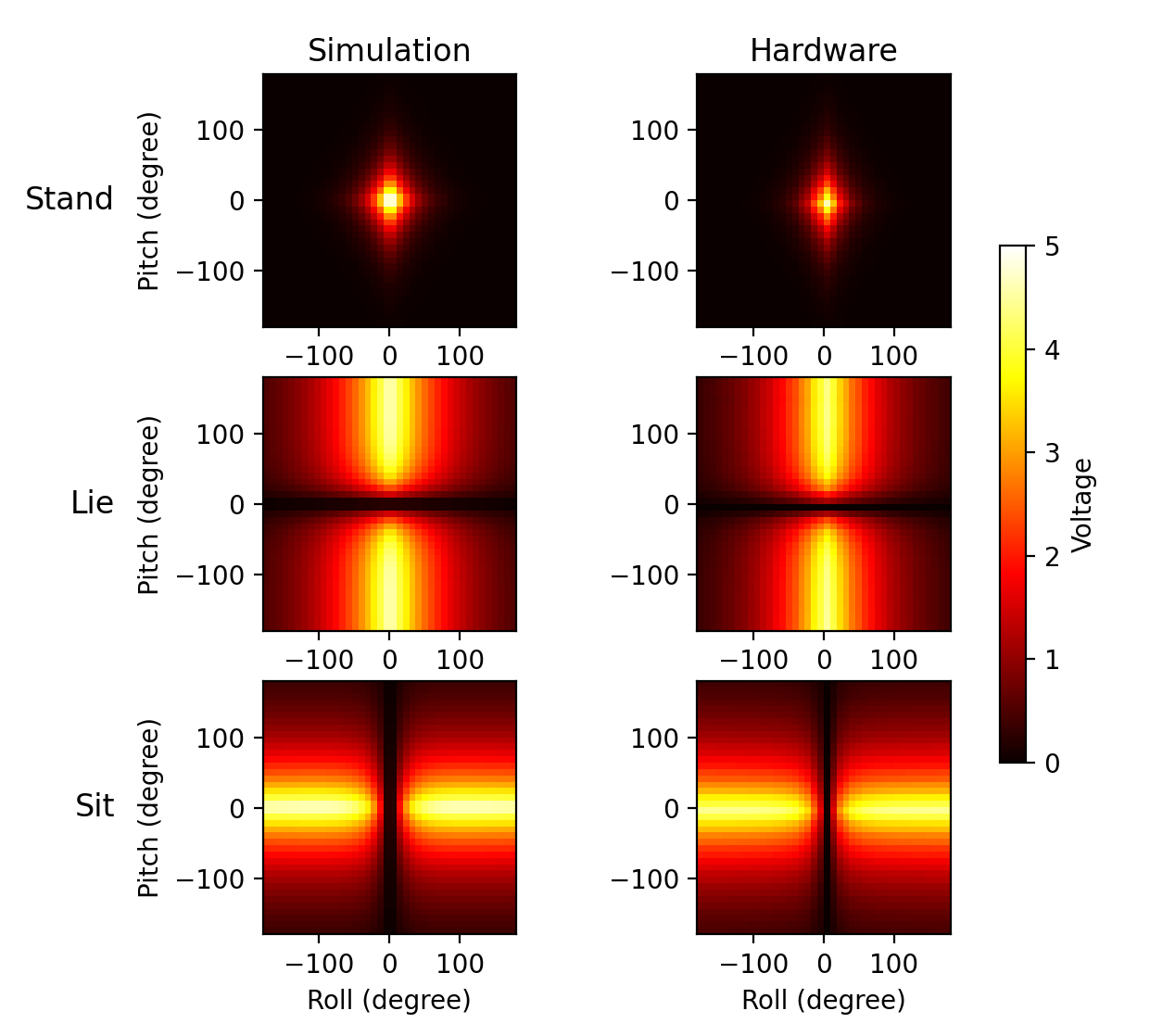}
    \caption{Comparison of neural responses between simulated and electronic IF neurons. Each simulated unit (i.e. Stand, Sit and Lie) is compared with its corresponding electronic implementation by measuring the capacitor voltage for all possible inputs (pitch and roll). The hardware implementation of the model is very close to the simulation behavior and only varies due to electric noise and rounded resistance values.}
    \label{fig:comparison}
\end{figure}

The model has been validated by sending all the possible inputs to the simulation and the hardware and comparing their responses. To achieve this, the hardware has not been tested using the inertial sensor but stimulated with the same tilt vectors as used in the simulation. Therefore, a mapping of the units' responses for both the simulation and the hardware has been generated – see Figure \ref{fig:comparison}. It appears that the behaviour of the electronic implementation is close to the simulation and the slight variations in voltage are due to noise and rounding of resistance values – e.g. a resistance of 3230$\Omega$ in the simulation is rounded to 3000$\Omega$ in the electronic implementation. Once the hardware is implemented and the model accuracy is validated, the power consumption of the device can be measured and compared to the use of simulated artificial neural networks.

\subsection{Power consumption analysis} \label{theorical_energy_consumpion}
\begin{table}
\caption{Comparison of average power consumptions of the micro-controller only, a Logistic Regression model running on the micro-controller and the micro-controller with the designed IF neurons. The values are given with and without the micro-controller power consumption to ease understanding.}
\label{power_consumption}
\begin{tabular}{|c|c|c|} 
\hline
Setup & Average power consumption & Average power consumption \\
& with the micro-controller & without the micro-controller \\
& (in Watt) & (in Watt) \\
\hline
\hline
Micro-controller only & 0.2155 & -\\ 
\hline
Logistic Regression & 0.2265 & 0.011 \\
on the micro-controller && \\
\hline
Micro-controller + IF circuits & \textbf{0.218} & \textbf{0.0025} \\
\hline
\end{tabular}
\end{table}
To measure the power efficiency of the hardware, the current consumed by the device (i.e. the micro-controller, inertial sensors and IF circuits together) has been recorded while performing real-time classification of dog postures. The measures were done using a power analyzer and power supply (Otii ARC) and performed on the micro-controller running alone, on the micro-controller classifying postures with the IF neurons implemented as hardware and on the micro-controller classifying postures with logistic regression. The following average power consumptions have been determined and written in Table \ref{power_consumption}. Most of the power consumed by the device is due to the micro-controller, but the results show that the use of the IF circuit for embedded classification consumes less than a simple logistic regression performed on the micro-controller. If the power consumption of the micro-controller is ignored, the implemented hardware consumes 4.4 times less than the logistic regression method, which is significant. Therefore, the designed circuit is faithful to its simulation and able to recognize patterns in presented inputs with less energy demand than simulated ANNs.

\section{Discussion} \label{section_discussion}

In this work, the Integrate-and-Fire model has been simulated and trained to achieve dog posture classification and showed promising results with relatively low energy expenditure when implemented with electronic components compared to the use of embedded logistic regression. The designed model implemented as hardware can be integrated into remote systems for embedded and energy-efficient pattern recognition, reducing the amount of data to transmit and thus reducing the number of transmissions, leading to low energy consumption.
The simulation of the IF neuron is faithful to the electronic implementation which makes it possible to train using the gradient descent algorithm. Once trained, the resistances that do not contribute to the pattern detection – i.e. those that converge to the highest value – are removed from the final circuit and the remaining are implemented with the final hardware. This hardware implementation has been done with only a few passive components (resistors, diodes and capacitors) and one active component (N-MOSFET transistors) which all have low costs. It has been implemented using prototyping boards but can be miniaturised on Printed Circuit Boards (PCBs) with Surface Mount Technology (SMT) that provides miniature components to produce a version of the hardware small enough to be integrated into small devices.
\par
In terms of power, the measured consumptions are almost identical due to the power demand of the micro-controller. However, the lifetime of battery-powered devices is very important and no aspect of the entire device should be overlooked, including the power usage of data processing. Therefore, by disregarding consumption of the micro-controller, the IF model consumes four times less when it is electronically implemented than a trained logistic regression running on the micro-controller. With this setup, the battery life-time is improved by 3.75\%, but it can be enhanced even more by using a low-powered micro-controller. Moreover, an implementation of the model with spike trains should significantly reduce energy consumption. Therefore, it would be wise to rethink the way of communicating features given to the model using spike trains to further reduce the power consumption of the circuit.
\par
One main issue of the IF approach is the time dependence of the inference. As the stimulation time of synapses varies according to the input values, the inference time is also variable. Thus, the higher the inputs, the longer the inference time will be. This maximum inference time can be calculated by summing the maximum stimulation time of inputs or can be compensated by varying the capacitance value of units. Another issue of the IF model is that the leak channel, specific to the LIF neuron, has been removed and the time dimension disappeared. This model is thus no longer able to process animal dynamics to infer its activity and only the posture – i.e. static patterns – can be classified. To achieve this task, the LIF model should be used which involves transforming features into spike trains. However, due to the non-continuity of spike trains in spiking neural networks, algorithms based on differentiation – such as the gradient descent algorithm used in this work – cannot be applied for training.

\section{Future work}
In future works, the time capabilities of the Leaky Integrate-and-Fire model must be exploited to classify time-series patterns using spike trains. As the gradient descent algorithm is not suitable to train such models, other training algorithms must be explored to find new ways to classify patterns or compress sensory data into a spike code generated by a spiking neural network. Recent advances in neurosciences permitted the development of unsupervised learning algorithms such as Spike Time Dependent Plasticity (STDP) which is a biologically plausible Hebbian learning rule that adjusts the strength of connections between neurons in the brain \cite{memristor_emulator_with_stdp,spike_time_dependence_of_plasticity,SNN_single_neurons_populations_plasticity}. Based on the timing of pre and post-synaptic spikes, STDP allows neurons to learn time-dependent correlations in spike trains and thus a relevant representation of input features \cite{memristor_emulator_with_stdp,spike_time_dependence_of_plasticity,SNN_single_neurons_populations_plasticity}. Therefore, such algorithms may be able to find correlations between some sensory inputs and achieve a compression of recorded data.

\bibliographystyle{splncs04}
\bibliography{bibliography.bib}

\end{document}